# Massively Distributed SGD: ImageNet/ResNet-50 Training in a Flash

**Hiroaki Mikami, Hisahiro Suganuma, Pongsakorn U-chupala,**
**Yoshiki Tanaka and Yuichi Kageyama**
Sony Corporation
{Hiroaki.Mikami, Hisahiro.Suganuma, Pongsakorn.Uchupala,
Yoshiki.Tanaka, Yuichi.Kageyama}@sony.com

## Abstract

Scaling the distributed deep learning to a massive GPU cluster level is challenging due to the instability of the large mini-batch training and the overhead of the gradient synchronization. We address the instability of the large mini-batch training with batch-size control and label smoothing. We address the overhead of the gradient synchronization with 2D-Torus all-reduce. Specifically, 2D-Torus all-reduce arranges GPUs in a logical 2D grid and performs a series of collective operation in different orientations. These two techniques are implemented with Neural Network Libraries (NNL)[1] . We have successfully trained ImageNet/ResNet-50 in 122 seconds without significant accuracy loss on ABCI[2] cluster.

## 1    Introduction

As the size of datasets and deep neural network (DNN) model for deep learning increase, the time required to train a model is also increasing. Large-scale distributed deep learning with data parallelism is an obvious course to effectively reduce the training time. However, there are two technical issues with large-scale distributed deep learning with a massive GPU cluster. The first issue is convergence accuracy degradation with large mini-batch training [1] [2]. The second issue is communication overhead of gradient synchronization among GPUs. A new approach to distributed processing is required to address these two issues.

In the past few years, many techniques have been proposed [1] [3] [4] [5] [6] to address these two issues. These works utilize ImageNet/ResNet-50 training to benchmark the training performance. ImageNet/ResNet-50 is one of the most popular datasets and DNN models for benchmarking large-scale distributed deep learning. Table 1 compares the training time and top-1 validation accuracy of the recent works.

The instability of a large mini-batch training and the gradient synchronization overhead are the primary issues that we addressed. Our best effort reduces the training time to 122 seconds with the validation accuracy of 75.29% using 3456 Tesla V100 GPUs. We also attempt to improve GPU scaling efficiency without significant accuracy loss. We achieved the GPU scaling efficiency 84.75% with 1024 Tesla V100 GPUs (Table 2).

---

[1]  An open source deep learning library, developed by Sony. https://nnabla.org/
[2]  AI Bridging Cloud Infrastructure (ABCI) is the world's first large-scale Open AI Computing Infrastructure, constructed and operated by National Institute of Advanced Industrial Science and Technology (AIST). https://abci.ai/

Table 1: Training time and top-1 1-crop validation accuracy with ImageNet/ResNet-50

|  | Batch Size | Processor | DL Library | Time | Accuracy |
|---|---|---|---|---|---|
| **He et al. [7]** | 256 | Tesla P100 x8 | Caffe | 29 hours | 75.3% |
| **Goyal et al. [1]** | 8K | Tesla P100 x256 | Caffe2 | 1 hour | 76.3% |
| **Smith et al. [4]** | 8K→16K | full TPU Pod | TensorFlow | 30 mins | 76.1% |
| **Akiba et al. [5]** | 32K | Tesla P100 x1024 | Chainer | 15 mins | 74.9% |
| **Jia et al. [6]** | 64K | Tesla P40 x2048 | TensorFlow | 6.6 mins | 75.8% |
| **Ying et al. [8]** | 32K | TPU v3 x1024 | TensorFlow | 2.2 mins | 76.3% |
| **Ying et al. [8]** | 64K | TPU v3 x1024 | TensorFlow | 1.8 mins | 75.2% |
| **This work** | **54K** | **Tesla V100 x3456** | **NNL** | **2.0 mins** | **75.29%** |

Table 2: GPU scaling efficiency with ImageNet/ResNet-50 training

|  | Processor | Interconnect | GPU scaling efficiency |
|---|---|---|---|
| **Goyal et al. [1]** | Tesla P100 x256 | 50Gbit Ethernet | ~90% |
| **Akiba et al. [5]** | Tesla P100 x1024 | Infiniband FDR | 80% |
| **Jia et al. [6]** | Tesla P40 x1024 | 100Gbit Ethernet | 87.9% |
| **This work** | **Tesla V100 x1024** | **Infiniband EDR x2** | **84.75%** |

## 2 Approach

There are two primary issues with large-scale distributed training: instability of large mini-batch training and the synchronization communication overhead.

It is well-known that training with large mini-batch is unstable and creates generalization gap [1] [2] [9]. In up to 32K mini-batch training on ImageNet/ResNet-50, this instability was alleviated by several groups [1] [5] [10]. Besides this, [6] has achieved training with 64K mini-batch.

A data parallel distributed training requires an extra step between every training iteration to synchronize and average gradients across participating GPUs. This step is implemented using an all-reduce collective operation. On a large-scale GPU cluster, the overhead of the all-reduce collective operation makes it extremely challenging to achieve linear scaling [5]

These two issues are addressed in this work. To address accuracy degradation, we tested two techniques: 1) batch-size control technique introduced in [4], [11], [12] and 2) label smoothing proposed in [13]. We develop 2D-Torus all-reducing scheme to efficiently exchange gradients across GPUs.

### 2.1 Techniques for Large Mini-batch Training

**Batch-Size Control:** According to the previous efforts, gradually increasing total mini-batch size during the training reduces the instability of the large mini-batch training. Intuitively, increasing the batch size as the loss landscape of the training becomes "flatter" helps evading the local minima [4] [11] [12]. In this work, we experimented with batch-size control to reduce accuracy degradation with mini-batch size exceeding 32K. A predetermined batch-size adjustment scheduling is employed during the training.

**Label Smoothing (LS):** Previous studies have shown that regularization can improve generalization [9] [6] [8]. We experimented with label-smoothing proposed in [13] as a regularization method to reduce accuracy degradation with mini-batch size exceeding 32K. Label smoothing decreases the probability value of the true label and increases the probability values of the false labels to avoid overfitting.

## 2.2 2D-Torus All-reduce

An efficient communication topology is vital for reducing communication overhead of a collective operation. Several communication topologies including Ring all-reduce [14] and hierarchical Ring all-reduce [6] are proposed to improve the efficiency of the all-reduce operation in the previous efforts.

Ring all-reduce algorithm cannot fully utilize the bandwidth of an extremely large-scale cluster with over thousand GPUs. This is because the communication overhead of the algorithm increases in proportion to the number of GPUs due to network latency as illustrated in [14].

We develop 2D-Torus all-reduce to address this problem. The 2D-Torus topology is described in Figure 1. The GPUs in the cluster are arranged in a 2D grid. In the 2D-torus topology, all-reduce consists of three steps: reduce-scatter, all-reduce, and all-gather. An example case of 2D-Torus all-reduce is shown in Figure 2. Firstly, reduce-scatter is performed horizontally. Then, all-reduce is performed vertically. Finally, all-gather is performed horizontally. Communication overhead of the 2D-Torus all-reduce is smaller than that of Ring all-reduce. Let $N$ be the number of GPUs in the cluster, $X$ be the number of GPUs in the horizontal direction, $Y$ be the number of GPUs in the vertical direction. 2D-Torus all-reduce executes $2(X-1)$ GPU-to-GPU operations. Comparatively, Ring all-reduce scheme executes $2(N-1)$ GPU-to-GPU operations [14]. While the hierarchical all-reduce also does the same amount of GPU-to-GPU operation as the 2D-Torus all-reduce, the data size of the second step (vertical all-reduce) of the 2D-Torus all-reduce scheme is $X$ times smaller than that of the hierarchical all-reduce.

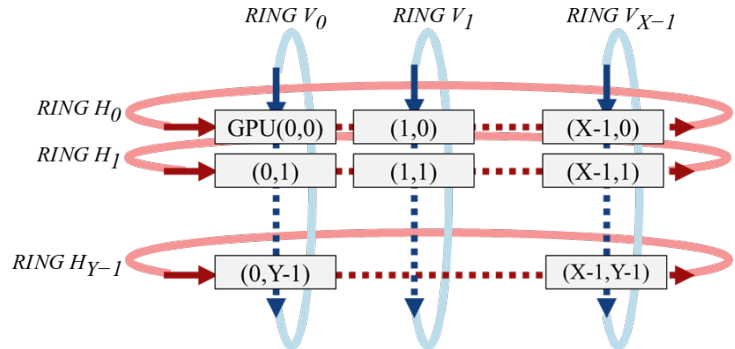

**Figure 1 : The 2D-Torus topology comprises of multiple rings in horizontal and vertical orientations.**

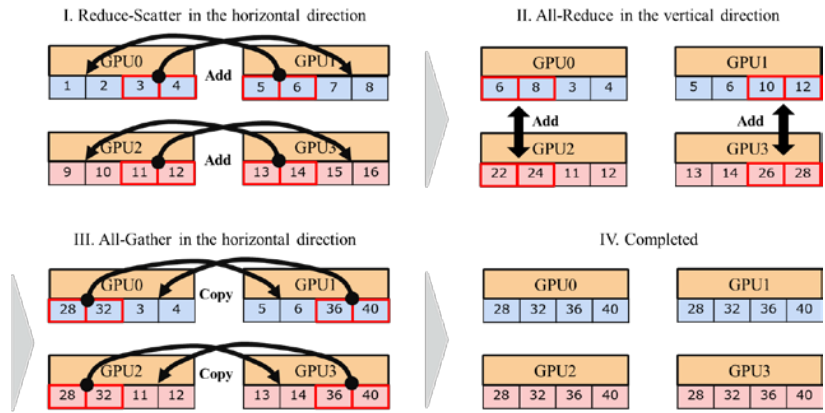

**Figure 2 : The 2D-Torus all-reduce steps of a 4-GPU cluster, arranged in 2x2 grid**

# 3 Evaluation
## 3.1 Experimental Environments

**Software:** We used Neural Network Libraries (NNL) and its CUDA extension as a DNN training framework. We used development branches based on NNL version 1.0.8. CUDA version 9.2 with cuDNN version 7.4.2 is employed to train DNN using multiple GPUs. We used a development branch based on NCCL version 2.4.0.a1 and OpenMPI version 2.1.3 as communication libraries. The 2D-Torus all-reduce is implemented with NCCL.

**Hardware:** We used AI Bridging Cloud Infrastructure (ABCI) as a GPU cluster. ABCI is a GPU cluster operated by National Institute of Advanced Industrial Science and Technology (AIST). It includes 1088 nodes. Each node has 4 NVIDIA Tesla V100 GPUs, 2 Xeon Gold 6148 processors, and 376 GB of memory. GPUs in the same node are connected with NVLink2 interconnect, whereas nodes are connected with 2 InfiniBand EDR interconnects.

## 3.2 Experimental Settings

**Dataset and Model:** We used ImageNet [15] dataset. This is a dataset for 1,000 classes image classification. ImageNet consists of 1.28 million training images and 50,000 validation images. We used NNL's implementation of image augmentation operations including padding, scaling, rotations, resizing, distortion, flipping, brightness adjustment, contrast adjustment, and noising in all our experiments. We used ResNet-50 [7] as a DNN model. All layers in the model are initialized by the values described in [10].

**Training Configurations:** We used LARS [10] with a coefficient of 0.01 and eps of 1e-6 (both are default configurations) to update the weights. We conducted preliminary experiments to optimize learning-rate (LR) and momentum. In this work, we employed two configurations: The first configuration (referred to as A) is obtained from the TensorFlow repository[3]. The second configuration (referred to as B) is based on the original settings reported in [10]. Both training configurations are shown in Table 3.

For configuration A, we used 34-epochs LR warmup, 34.0 as the base LR, and $10^{-5}$ as the initial LR. The momentum is 0.9. For configuration B, the learning-rate (LR) and the momentum are calculated with the following formula. We used 5-epochs LR warmup. The base LRs of 29 and 50 are the exact value used in [10] and the maximum value suggested in [3] respectively. We used the relation between the total mini-batch size, LR, and momentum proposed by [16] to calculate the momentum.

$$epoch = \frac{ProcessedSamples}{DataSize}$$

$$LearningRate(epoch) = \begin{cases} 0.2 + (29-0.2)\frac{epoch}{5} & \text{if } epoch < 5 \\ 29\left(1 - \frac{epoch}{90}\right)^2 & \text{if } epoch < 30 \\ 50\left(1 - \frac{epoch}{90}\right)^2 & \text{otherwise} \end{cases}$$

$$NoiseScale(epoch) = \frac{LearningRate(epoch) \cdot 32 \cdot 1024}{1-0.9}$$

$$Momentum(epoch) = 1 - \frac{LearningRate(epoch) \cdot B}{NoiseScale(epoch)}$$

---

[3] https://github.com/tensorflow/tpu/blob/f4fa4d7c167040d6541ac08d0aae44bc7ee13609/models/official/resnet/lars_util.py

Table 3: Training configurations used in our experiments

|  | #GPUs (Max) | LS | LR | Mini-batch Size | | | |
|---|---|---|---|---|---|---|---|
|  |  |  |  | Epoch 1-30 | Epoch -45 | Epoch -75 | Epoch -90 |
| Reference[4] | 1024 |  | - | 32 / worker Total 32K |  |  |  |
| Exp. 1[5] | 2176 |  | A | 16 / worker Total 34K | 32 / worker Total 68K |  |  |
| Exp. 2 | 3456 | ✓ | B | 16 / worker Total 54K | 32 / worker Total 54K |  |  |
| Exp. 3 | 3456 | ✓ | B | 16 / worker Total 54K | 32 / worker Total 64K |  |  |
| Exp. 4 | 4096 | ✓ | A | 16 / worker Total 34K | 16 / worker Total 68K | 32 / worker Total 85K | 32 / worker Total 119K |

Table 4: The grid dimensions of the 2D-Torus topology used in our experiments

| #GPUs | Vertical | Horizontal |
|---|---|---|
| 1024 | 32 | 32 |
| 2048 | 32 | 64 |
| 2176 | 34 | 64 |
| 3456 | 48 | 72 |
| 4096 | 64 | 64 |

We also employed mixed-precision training introduced in [17]. The forward/backward computations and the communication to synchronize gradients are conducted in half precision float (FP16). The computation in LARS was conducted in single precision float (FP32) because LARS required a wider dynamic range than the FP16 format [6]. We also employed "Batch Normalization without Moving Average" [5] to get the accurate sample mean and variance. The communication to synchronize batch mean and batch squared mean was also conducted in FP32 due to the wider dynamic range.

### 3.3 Results

We finished the ResNet-50 training in 122 seconds with no significant accuracy loss as shown in Table 5. Both batch-size controlling and label smoothing help to enable large mini-batch training. Using only batch-size control technique (Exp. 4 in Table 5), the maximum mini-batch size can be increased up to 119K with no significant accuracy loss. Using only label smoothing technique (Exp. 2 in Table 5), the initial mini-batch size can be increased up to 54K with no significant accuracy loss. However, combining both techniques with large mini-batch size (Exp. 3 in Table 5) decreases the accuracy by about 0.7%.

Although utilizing faster GPUs (Tesla V100), the 2D-Torus communication scheme achieved GPU scaling efficiency to the previous research [6]. The previous research achieved the GPU scaling efficiency of 87.9% using 1024 Tesla P40 GPUs with per-worker mini-batch size set to 32. Table 6 shows the number of GPUs and training throughput of 2D-Torus communication scheme with per-worker mini-batch size set to 32. The GPU scaling efficiency, relative to the single-node (4 GPUs) performance, exceeds 80% with 1024 and 2048 GPUs and scales reasonably well up to 4096 GPUs.

## 4 Conclusion

Large-scale distributed deep learning is an effective approach to reduce a DNN training time. We employ several techniques to reduce accuracy degradation while maintaining high GPU scaling efficiency when training with an enormous GPU cluster. The techniques are implemented using Neural Network Libraries. We achieved the training time of 122 seconds and the validation accuracy of 75.29% using 3456 Tesla V100 GPUs. We also achieved over 80% GPU scaling efficiency with up to 2048 Tesla V100 GPUs.

---

[4] Training settings (e.g., hyper parameters) reported in [10] are used.
[5] This experiment was conducted with previous software version.

Table 5: Top-1 1-crop validation accuracy and training time

|  | #GPUs (Max) | Batch Size (Min/Max) | Validation Accuracy | Time |
|---|---|---|---|---|
| Reference | 1024 | 32K | 75.40% | 505 secs |
| Exp. 1 | 2176 | 34K/68K | 75.03% | 224 secs |
| Exp. 2 | 3456 | 54K | 75.29% | 122 secs |
| Exp. 3 | 3456 | 54K/64K | 74.62% | 115 secs |
| Exp. 4 | 4096 | 34K/119K | 75.23% | 129 secs |

Table 6: Training throughput and scaling efficiency of 2D-Torus communication scheme with per-worker mini-batch size set to 32

| #GPUs | Images per Second | GPU Scaling Efficiency |
|---|---|---|
| 4 | 2565 | - |
| 1024 | 556522 | 84.75% |
| 2048 | 1091357 | 83.10% |
| 3456 | 1641853 | 74.08% |
| 4096 | 1929054 | 73.44% |

**Acknowledgments**

Computational resource of AI Bridging Cloud Infrastructure (ABCI) was awarded by "ABCI Grand Challenge" Program, National Institute of Advanced Industrial Science and Technology (AIST).

The authors would like to thank K. Yoshiyama, T. Narihira, Y. Kobayashi and A. Nakamura for the technical advice as well as A. Shin for the help regarding the manuscript.